%% file: root.tex
\documentclass[conference]{IEEEtran}
\IEEEoverridecommandlockouts
\usepackage{cite}
\usepackage{amsmath,amssymb,amsfonts}
\usepackage{graphicx}
\usepackage{textcomp}
\usepackage{xcolor}
\def\BibTeX{{\rm B\kern-.05em{\sc i\kern-.025em b}\kern-.08em
    T\kern-.1667em\lower.7ex\hbox{E}\kern-.125emX}}

\usepackage[utf8]{inputenc} 
\usepackage[T1]{fontenc}    
\usepackage{hyperref}       
\usepackage{url}            
\usepackage{booktabs}       
\usepackage{amsfonts}       
\usepackage{nicefrac}       
\usepackage{microtype}      
\usepackage{lipsum}		
\usepackage{graphicx}
\usepackage{subcaption}
\usepackage{graphicx}
\usepackage{xcolor} 
\usepackage{tikz}
\usetikzlibrary{shapes.geometric, arrows, positioning, automata,positioning,fit,backgrounds}
\tikzstyle{process} = [rectangle, minimum width=2cm, minimum height=1cm, text centered, draw=black, fill=white!30]
\tikzstyle{sum} = \tikzstyle{sum} = [draw, circle, minimum size=.5cm]
\tikzstyle{arrow} = [thick,->,>=stealth]

\usepackage{amsmath}

\usepackage{bm}
\usepackage{units}
\usepackage{float}
\usepackage{dblfloatfix}
\usepackage{algorithm}
\usepackage{algpseudocode}
\makeatother
\usepackage[T1]{fontenc}

\usetikzlibrary{fit,calc}
\newcommand*{\tikzmk}[1]{\tikz[remember picture,overlay,] \node (#1) {};\ignorespaces}
\newcommand{\boxit}[1]{\tikz[remember picture,overlay]{\node[yshift=0pt,fill=#1,opacity=.25,fit={(A)($(B)+(.9\linewidth,.5\baselineskip)$)}] {};}\ignorespaces}
\colorlet{yellow}{yellow!100}
\colorlet{blue}{cyan!60}

\begin{document}

\title{Exploring Robot Morphology Spaces through Breadth-First Search and Random Query}


\author{\IEEEauthorblockN{Jie Luo}
\IEEEauthorblockA{\textit{Computer Science Department} \\
\textit{Vrije Universiteit Amsterdam}\\
Amsterdam, The Netherlands \\
j2.luo@vu.nl}
}

\maketitle

\begin{abstract}

Evolutionary robotics offers a powerful framework for designing and evolving robot morphologies, particularly in the context of modular robots. However, the role of query mechanisms during the genotype-to-phenotype mapping process has been largely overlooked. This research addresses this gap by conducting a comparative analysis of query mechanisms in the brain-body co-evolution of modular robots. Using two different query mechanisms, Breadth-First Search (BFS) and Random Query, within the context of evolving robot morphologies using CPPNs and robot controllers using tensors, and testing them in two evolutionary frameworks, Lamarckian and Darwinian systems, this study investigates their influence on evolutionary outcomes and performance. The findings demonstrate the impact of the two query mechanisms on the evolution and performance of modular robot bodies, including morphological intelligence, diversity, and morphological traits. This study suggests that BFS is both more effective and efficient in producing highly performing robots. It also reveals that initially, robot diversity was higher with BFS compared to Random Query, but in the Lamarckian system, it declines faster, converging to superior designs, while in the Darwinian system, BFS led to higher end-process diversity.

\end{abstract}


\section{Introduction}
Evolutionary robotics empowers the design and evolution of robot morphologies through a process of genotype to phenotype mapping. In the context of modular robots, the challenge lies in determining the presence or absence of specific components at precise positions within the robot body and getting a balance in exploring and exploiting the design space.

Several genotype-to-phenotype mapping techniques have been employed in various research studies, including L-systems \cite{miras2018effects}, CPPNs (Compositional pattern-producing networks) \cite{Stanley2007,Luo2022,lipson2016difficulty}, and Direct Mapping \cite{Collins2019}. However, scant attention has been given to the query mechanism utilized in these mapping processes, despite its pivotal role in shaping the resultant robot bodies.

This research aims to address the open research area of investigating different query mechanisms in the evolutionary robotic field. The primary objective is to conduct a comparative analysis of query mechanisms and their influence on the evolution and performance of modular robot bodies. These investigations focus on understanding how different query mechanisms affect the key characteristics of evolved robot morphologies in evolutionary robot systems.

To achieve this objective, we design and implement an experimental setup where we evolve modular robot morphologies using CPPNs with one commonly used query mechanism: Breadth-First Search (BFS) \cite{bundy1984breadth} and compare it with our design: Random Query \cite{Miras2023}. We test these two query mechanisms on two evolutionary systems to evolve both the body and brain. 

The main contributions of this research are threefold. Firstly, we provide a comprehensive analysis of the influence of two different query mechanisms on the evolution and performance of modular robot morphologies. 

Secondly, we contribute to the understanding of genotype to phenotype mapping in modular robotics by highlighting the importance of the query mechanism and its impact on the diversity and complexity of evolved robot morphologies. Our findings can inform the development of more effective approaches for evolving robot bodies and contribute to the advancement of adaptive and versatile robotic systems.

Finally, we evaluate the efficiency and convergence properties of the query mechanisms, considering the computational resources required for generating desirable robot body configurations. This analysis provides valuable insights for researchers and practitioners working on evolutionary robotics, enabling them to make informed decisions regarding the choice of query mechanism based on their specific requirements and constraints.

Overall, this research enhances our understanding of query mechanisms in genotype to phenotype mapping for modular robots and sheds light on key aspects of evolutionary robotics. 

\section{Evolution+Learning}
A search space comprises distinct layers that stack upon one another. At its foundational level lies the phenotype space, while one layer above that resides the genotype space, which may not always have a straightforward one-to-one representation with the phenotype layer. Numerous factors influence our search process, including reproduction operators and selection mechanisms, among others. Our particular focus revolves around examining how the query mechanisms employed for mapping the body genotype to the robot's morphology impact the exploration of the morphological search space.

\subsection{Robot Phenotype}

\subsubsection{Robot Morphology}

We adopt RoboGen's components as the robot body's phenotype. RoboGen \cite{Auerbach2014} is a popular open-source platform for evolving robots, offering modular components: a core component, one or more brick components, and active hinges. The phenotype follows a tree structure, with the core module as the root node, enabling 3D morphologies through 90-degree rotations.

\subsubsection{Robot Controller}

We employ Central Pattern Generators (CPGs) for driving modular robots, a proven method for controlling various robot types \cite{Luo2022,lan2021learning}. Each robot joint has an associated CPG consisting of three neurons: an $x_i$-neuron, a $y_i$-neuron, and an $out_i$-neuron. The $x_i$ and $y_i$ neuron states change over time by multiplying the activation value of the opposing neuron by a corresponding weight: $\dot{x}_i = w_i y_i$ and $\dot{y}_i = -w_i x_i$. To simplify, we set $w_{x_iy_i}$ equal to $-w_{y_ix_i}$, denoting their absolute value as $w_i$. Initial states of all $x$ and $y$ neurons are $\frac{\sqrt{2}}{2}$ to create a sine wave with an amplitude of 1, matching joint rotation limits.

To allow complex output patterns, we implement connections between neighboring joint CPGs. For the $i_{th}$ joint and $\mathcal{N}_i$ as the set of neighboring joint indices, with $w_{ij}$ representing the connection weight between $x_i$ and $x_j$ (also set to $-w_{ji}$), the system of differential equations becomes:

\begin{equation}
    \begin{split}
        \dot{x}_i &= w_i y_i + \sum_{j \in \mathcal{N}_i} w_{ji} x_j \\
        \dot{y}_i &= -w_i x_i
    \end{split}
\end{equation}

Due to this addition, $x$ neurons are no longer bounded within $[-1,1]$. To handle this, we use the hyperbolic tangent function (\emph{tanh}) as the activation function for $out_i$-neurons.


\begin{figure}
   \begin{minipage}{0.49\textwidth}
     \centering
     \includegraphics[width=.9\linewidth]{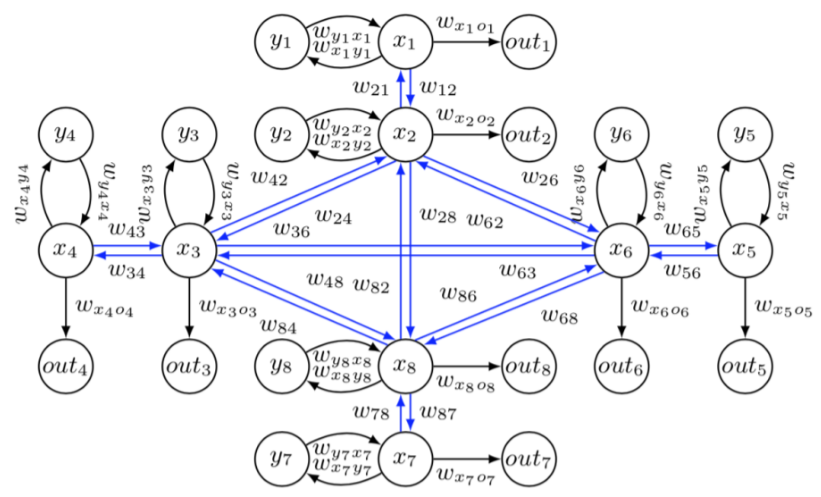}
    \end{minipage}
   \centering
    \caption{\label{fig:cpg_network}Brain phenotype (CPG network) of a "+" shape robot. In our design, the topology of the brain is determined by the topology of the body.}
\end{figure}
 \vspace{-5pt}

\subsection{Robot Genotype}

\subsubsection{Body Genotype}
The phenotype of bodies is encoded in a Compositional Pattern Producing Network (CPPN) which was introduced by Stanley \cite{Stanley2007} and has been successfully applied to the evolution of both 2D and 3D robot morphologies in prior studies \cite{luo2023}. The structure of the CPPN has four inputs and five outputs. The first three inputs are the x, y, and z coordinates of a component, and the fourth input is the distance from that component to the core component in the tree structure. The first three outputs are the probabilities of the modules being a brick, a joint, or empty space, and the last two outputs are the probabilities of the module being rotated 0 or 90 degrees. For both module type and rotation the output with the highest probability is always chosen; randomness is not involved.


\subsubsection{Brain Genotype}
We utilize an array-based structure for the brain's genotypic representation to map the CPG weights. This is achieved via direct encoding, a method chosen specifically for its potential to enable reversible encoding in future stages.
We have seen how every modular robot can be represented as a 3D grid in which the core module occupies the central position and each module's position is given by a triple of coordinates. When building the controller from our genotype, we use the coordinates of the joints in the grid to locate the corresponding CPG weight. To reduce the size of our genotype, instead of the 3D grid, we use a simplified 3D in which the third dimension is removed. For this reason, some joints might end up with the same coordinates and will be dealt with accordingly. 

Since our robots have a maximum of 10 modules, every robot configuration can be represented in a grid of $21 \times 21$. Each joint in a robot can occupy any position of the grid except the center. For this reason, the possible positions of a joint in our morphologies are exactly $(21 \cdot 21) - 1=440$. We can represent all the internal weights of every possible CPG in our morphologies as a $440$-long array. When building the phenotype from this array, we can simply retrieve the corresponding weight starting from a joint's coordinates in the body grid.

To represent the external connections between CPGs, we need to consider all the possible neighbours a joint can have. In the 2-dimensional grid, the number of cells in a distance-2 neighbourhood for each position is represented by the Delannoy number $D(2,2) = 13$, including the central element. Each one of the neighbours can be identified using the relative position from the joint taken into consideration. Since our robots can assume a 3D position, we need to consider an additional connection for modules with the same 2D coordinates.

\begin{figure}
    \centering
    \includegraphics[width=0.47\textwidth]{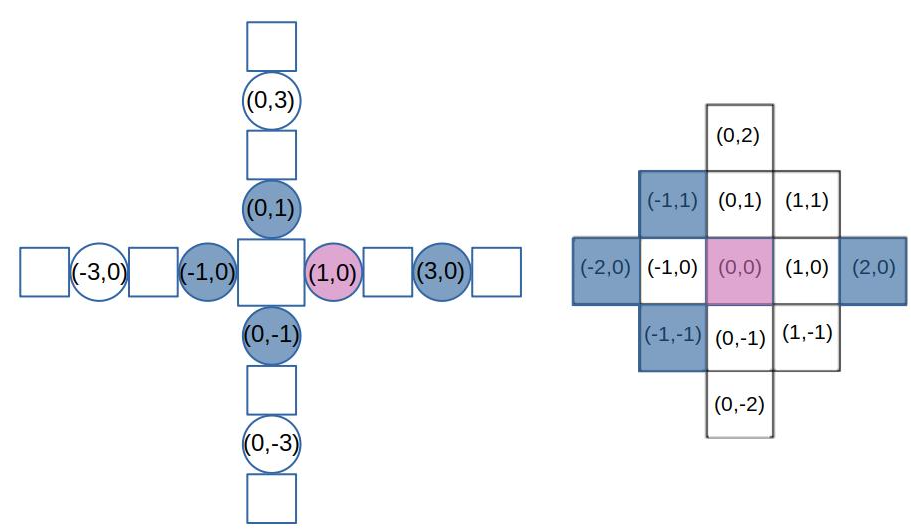}
    \caption{Brain genotype to phenotype mapping of a "+" shape robot. The left image (brain phenotype)  shows the schema of the "+" shape robot with the coordinates of its joints in the 2D body grid. The right image (brain genotype) is the distance 2 neighbour of the joint at (1,0). The coordinates reported in the neighbourhood are relative to this joint. The CPG weight of the joint is highlighted in purple and its 2-distance neighbours are in blue.}
    \label{fig:brain_geno}
\end{figure}

To conclude, for each of the $440$ possible joints in the body grid, we need to store 1 internal weight for its CPG, 12 weights for external connections, and 1 weight for connections with CPGs at the same coordinate for a total of 14 weights. The genotype used to represent the robots' brains is an array of size $440 \times 14$. An example of the brain genotype of a "+" shape robot is shown in Figure \ref{fig:brain_geno}.

It is important to notice that not all the elements of the genotype matrix are going to be used by each robot. This means that their brain's genotype can carry additional information that could be exploited by their children with different morphologies.

\subsection{Query Mechanisms}
Query Mechanism is a critical aspect of the genotype-to-phenotype translation process in designing robot bodies. It serves as the bridge between the genetic information encoded in the genotypes (such as CPPN, L-system, array) and the actual physical characteristics of the robot. Essentially, the query mechanism is a technique used to extract information from the genotypic representation to determine the composition and arrangement of modules in the resulting robot body.

To produce the phenotypes of the robot bodies, the core component is generated at the origin. Then, two different mechanisms are used to query the CPPN-based genotypes:

\paragraph{Breadth-First Search} an algorithm for searching a tree data structure for a node that satisfies a given property \cite{bundy1984}. It starts at the tree root and explores all nodes at the present depth prior to moving on to the nodes at the next depth level. We move outwards from the core component until there are no open sockets(breadth-first exploration), querying the CPPN network to determine whether a module will be placed at each location, its type and its rotation. If a module would be placed in a location already occupied by a previous module, the module is simply not placed and the branch ends there.

\paragraph{Random Query} an algorithm for searching a tree data structure for a node randomly with a given number of queries. All open sockets have an equal chance of being randomly selected to be queried, in no specific order. The CPPN network determines the type and rotation of each module. If a module would be placed in a location already occupied by a previous module, then this module is not expressed in the body. A number of nine queries are applied.

For both methods, the coordinates of each module are integers; a module attached to the front of the core module will have coordinates (0,1,0). We stop when ten modules have been created.




\begin{algorithm}[h!]
  \caption{Evolution+Learning}
  \label{alg:EL}
  \begin{algorithmic}[1]
    \State INITIALIZE robot population 
    \State EVALUATE each robot 
    \While{not STOP-EVOLUTION}
        \State SELECT parents; 
        \State RECOMBINE+MUTATE parents' bodies; 
        \State MUTATE parents' brains; 
        \State CREATE offspring robot body; 
        \State CREATE offspring robot brain; 
        
        \State INITIALIZE brain(s) for the learning process; 
        \While{not STOP-LEARNING}
            \State ASSESS offspring; 
            \State GENERATE new brain for offspring;
        \EndWhile 
        \State EVALUATE offspring with the learned brain; 
        
        \tikzmk{A}
        \State UPDATE brain genotype 

        \tikzmk{B} \boxit{yellow}
        \State SELECT survivors / UPDATE population
          
    \EndWhile
 \end{algorithmic}
\end{algorithm}

\subsection{Learning Algorithm}
We use Reversible Differential Evolution (RevDE) \cite{Tomczak2020} as the learning algorithm because it has proven to be effective in previous research \cite{Luo2022}. This method works as follows:
\begin{enumerate}
    \item Initialize a population with \textit{$\mu$} samples ($n$-dimensional vectors), $\mathcal{P}_{\mu}$. 
    \item Evaluate all \textit{$\mu$} samples.
    \item Apply the reversible differential mutation operator and the uniform crossover operator.\\
    \textit{The reversible differential mutation operator}: Three new candidates are generated by randomly picking a triplet from the population, $(\mathbf{w}_i,\mathbf{w}_j,\mathbf{w}_k)\in \mathcal{P}_{\mu}$, then all three individuals are perturbed by adding a scaled difference. 
    \item Perform a selection over the population based on the fitness value and select \textit{$\mu$} samples.
    \item Repeat from step (2) until the maximum number of iterations is reached.
\end{enumerate}

As explained above, we apply RevDE here as a learning method for `newborn' robots. In particular, it will be used to optimize the weights of the CPGs of our modular robots for the tasks during the Infancy stage. 

The Algorithm \ref{alg:EL} displays the pseudocode of the complete integrated process of evolution and learning. With the highlighted yellow code, it is the Lamarckian system, without it is the Darwinian system. Note that for the sake of generality, we distinguish two types of quality testing depending on the context, evolution or learning.  

\subsection{Task and Fitness function}
Point navigation is a closed-loop controller task which needs feedback (coordinates)from the environment passing to the controller to steer the robot. The coordinates are used to obtain the angle between the current position and the target. If the target is on the right, the right joints are slowed down and vice versa. 

A robot is spawned at the centre of a flat arena (10 × 10 m\textsuperscript{2}) to reach a sequence of target points $P_1,..., P_N$. In each evaluation, the robot has to reach as many targets in order as possible. Success in this task requires the ability to move fast to reach one target and then quickly change direction to another target in a short duration. A target point is considered to be reached if the robot gets within 0.01 meters from it. Considering the experimental time, we set the simulation time per evaluation to be 40 seconds which allows robots to reach at least 2 targets $P_1(1,-1), P_2(0,-2)$.

The data collected from the simulator is the following:
\begin{itemize}
    \item The coordinates of the core component of the robot at the start of the simulation are approximate to $P_0 (0,0)$;
    \item The coordinates of the robot, sampled during the simulation at 5Hz, allowing us to plot and approximate the length of the followed path;
    \item The coordinates of the robot at the end of the simulation $P_T(x_T,y_T)$;
    \item The coordinates of the target points $P_1(x_1,y_1)$... $P_n(x_n,y_n)$.
    \item The coordinates of the robot, sampled during the simulation at 5Hz, allow us to plot and approximate the length of the path $L$.
\end{itemize}

The fitness function for this task is designed to maximize the number of targets reached and minimize the path followed by the robot to reach the targets.
 \vspace{-5pt}
\begin{multline}
    F=\sum_{i=1}^{k}dist(P_i,P_{i-1}) \\
    +(dist(P_k,P_{k-1}) - dist(P_T,P_k)) \\
    - \omega \cdot L
\end{multline}
where $k$ is the number of target points reached by the robot at the end of the evaluation, and $L$ is the path travelled. The first term of the function is a sum of the distances between the target points the robot has reached. The second term is necessary when the robot has not reached all the targets and it calculates the distance travelled toward the next unreached target. The last term is used to penalize longer paths and $\omega$ is a constant scalar that is set to 0.1 in the experiments. E.g., if a robot just reached 2 targets, the maximum fitness value will be $dist(P_1, P_0)+(dist(P_2, P_1)-dist(P2, P2))-0.1*L=\sqrt{2}+\sqrt{2}-0.2*\sqrt{2} \approx 2.54$ ($L$ is shortest path length to go through $P_1$ and $P_2$ which is equal to $2*\sqrt{2}$).

\section{Experimental Setup}
The stochastic nature of evolutionary algorithms requires multiple runs under the same conditions and a sound statistical analysis (\cite{bartz2007experimental}). We perform 10 runs for each query mechanism and evolutionary system, namely BFS Darwinian, BFS Lamarckian, Random Query Darwinian, and Random Query Lamarckian. In total, 40 experiments.

Each experiment consists of 30 generations with a population size of 50 individuals and 25 offspring. A total of $50+(25\cdot(30-1))=775$ morphologies and controllers are generated, and then the learning algorithm RevDE is applied to each controller. For RevDE we use a population of 10 controllers for 10 generations, for a total of $(10+30\cdot(10-1))=280$ performance assessments.

The fitness measures used to guide the evolutionary process are the same as the performance measure used in the learning loop. For this reason, we use the same test process for both.
The tests for the task of point navigation use 40 seconds of evaluation time with two target points at the coordinates of $(1, -1)$ and $(0, -2)$. 

All the experiments are run with Mujoco simulator-based wrapper called Revolve2 on a 64-core Linux computer, where they each take approximately 7 hours to finish.

The code for replicating this work and carrying out the experiments is available online: \url{https://shorturl.at/aES26}.

\begin{table}[h]
\caption{Main experiment parameters}
\begin{tabular}{{p{0.22\linewidth} | p{0.06\linewidth}| p{0.58\linewidth}}}
\toprule
Parameters       & Value & Description                                    \\ \midrule
Population size  & ~50    & Number of individuals per generation     \\
Offspring size  & ~25    & Number of offspring produced per generation     \\
Generations      & ~30   & Termination condition for each run             \\ 
Learning trials  & ~280    & Number of the evaluations performed by RevDE on each robot \\ 
Tournament size  & ~2     & Number of individuals used in the parent selection - (k-tournament)		 \\ 
Repetitions      &  ~10    & Number of repetitions per experiment \\ 
\bottomrule 
\end{tabular}
\label{tab:parameters}
\end{table}

\section{Results}
To compare the effects of BFS and Random Query, we consider two generic performance indicators: efficiency and efficacy, meanwhile we also look into robots' morphologies.
\subsection{Robot Performance}

\subsubsection{Efficacy} the average fitness in the final generation.
Figure \ref{fig:fitness_mean_avg} shows that both query mechanisms can produce robots able to solve the task, but robots queried by BFS are approximately 20\% better. Moreover, around generation 14, Lamarckian system had already significantly outperformed the result that was produced by Darwinian system only by the end of the evolutionary process. This holds true for both query mechanisms. 

\subsubsection{Efficiency} how much effort is needed to reach a given quality threshold (fitness level). It is calculated as the number of solution evaluations until the quality threshold is reached. 

\begin{figure*}[ht!] 
  \centering
     \begin{subfigure}[b]{0.47\textwidth}
         \centering
         \includegraphics[width=0.8\textwidth]{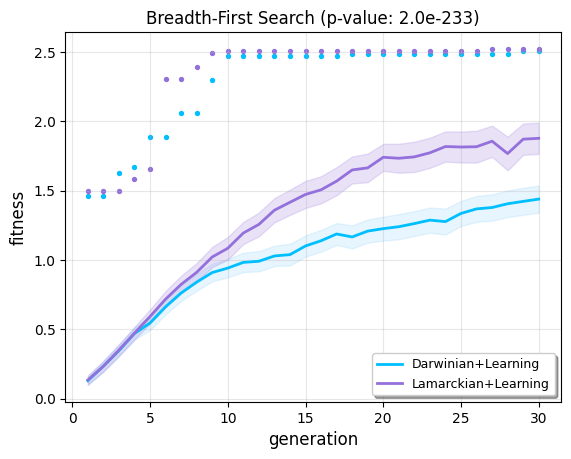}
         \caption{}
     \end{subfigure}
     \hfill
     \begin{subfigure}[b]{0.47\textwidth}
         \centering
         \includegraphics[width=0.8\textwidth]{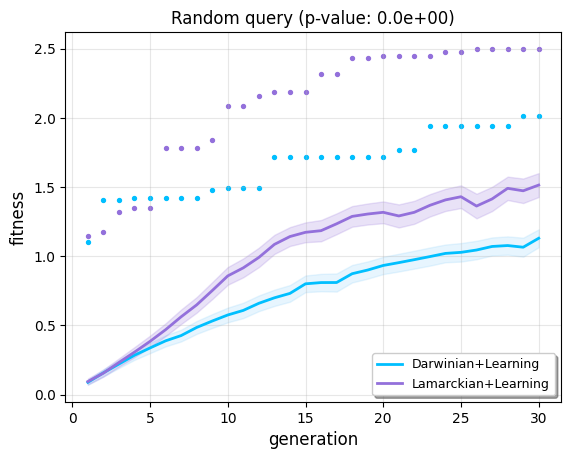}
         \caption{}
     \end{subfigure}
      \vspace{-5pt}
  \caption{Mean (lines) and maximum (dots) fitness over 30 generations (averaged over 10 runs) for Lamarckian system in purple and Darwinian system in blue. Subfigure (a) exhibits mean average fitness for robots produced with BFS, and Subfigure (b) is for Random Query. The bands indicate the 95\% confidence intervals ($\pm1.96\times SE$, Standard Error).}
  \label{fig:fitness_mean_avg} 
\end{figure*}

BFS in the Lamarckian system is the most efficient, as it finds the best solution (maximum fitness) fastest (Figure \ref{fig:fitness_mean_avg}). 

\subsection{Robot Morphologies}

\subsubsection{Morphological intelligence}
\begin{figure*}
    \centering
    \begin{subfigure}[b]{0.48\textwidth}
        \centering
\includegraphics[width=0.8\textwidth]{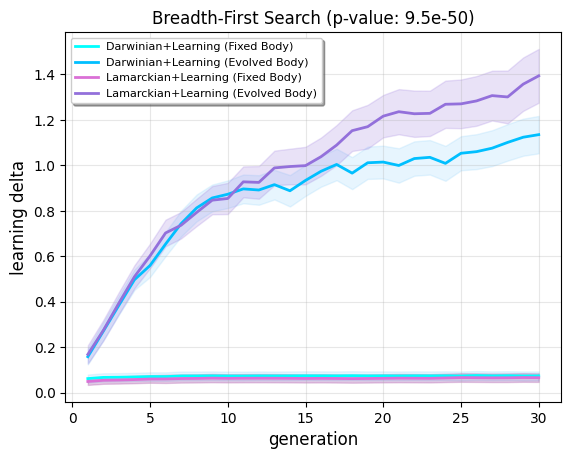}
    \end{subfigure}
    \hfill
    \begin{subfigure}[b]{0.48\textwidth}  
        \centering 
        \includegraphics[width=0.8\textwidth]{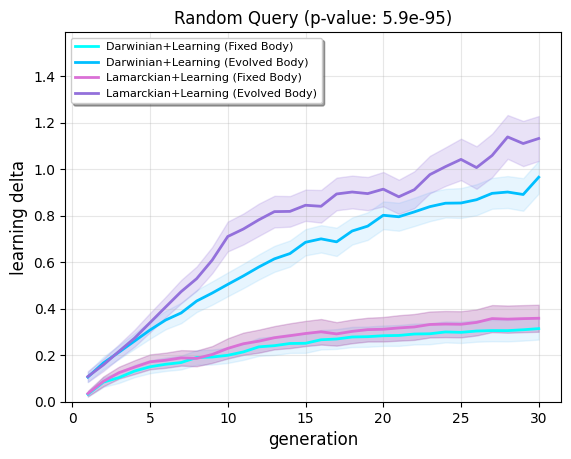}
    \end{subfigure}
    \caption{Progression of the learning $\Delta$, the difference between the robot's performance with its learned brain and the performance with its inherited brain, throughout evolution averaged over 10 runs. The P-value is shown in the title of the plot. The bands indicate the 95\% confidence intervals. The plot illustrates that the delta of learning delta between evolved body and fixed body for each method is growing across generations.}
    \label{fig:learning delta}
\end{figure*}

in this research, we consider a special property of robot morphology: Morphological Intelligence. Morphology influences how the brain learns. Some bodies are more suitable for the brains to learn with than others. How well the brain learns can be empowered by a better body. Therefore we define the intelligence of a body as a measure of how well it facilitates the brain to learn and achieve tasks. To quantify the measurement, we did an extra experiment, using the fixed bodies of 50 initial robots from the first generation of each run to evolve only the brains of them with these two methods, then we calculate the learning delta of each experiment, being the fitness value after the parameters were learned minus the fitness value before the parameters were learned. We finally quantify morphological intelligence by the delta of the learning delta of each method, being the learning delta of the evolved body minus the learning delta of the fixed body. In Figure \ref{fig:learning delta}, we see that the average learning $\Delta$ of both methods with evolved bodies grow steadily across the generations. This effect has been discovered previously in \cite{Miras2020, Luo2022}, with different tasks, a different learning method and a different representation, so the current results provide additional support that lifetime learning leads the evolutionary search towards morphologies with increasing learning potential. While the average learning Deltas of both methods with fixed body show no significant change which indicates that there is low morphological intelligence in the fixed robot body. The morphological intelligence in Lamarckian system is 30\% greater than that in Darwinian system, as indicated by the higher delta of the learning delta. The delta of learning delta in BFS is about 75\% higher than in Random Query, which indicates more morphological intelligence in the bodies produced by BFS.

\subsubsection{Diversity} the morphological variety of each population using tree-edit distance. 
It is measured in two steps: firstly, the measure of difference between any two robots, denoted as d(x,y); and secondly, the measure of diversity within a population, which is represented by the average distance along the evolutionary process.

\begin{figure*}[ht!] 
  \centering
     \begin{subfigure}[b]{0.48\textwidth}
         \centering
         \includegraphics[width=0.8\textwidth]{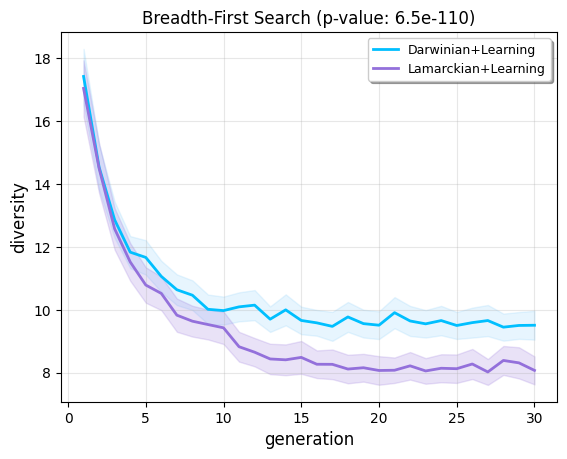}
         \caption{}
     \end{subfigure}
     \hfill
     \begin{subfigure}[b]{0.48\textwidth}
         \centering
         \includegraphics[width=0.8\textwidth]{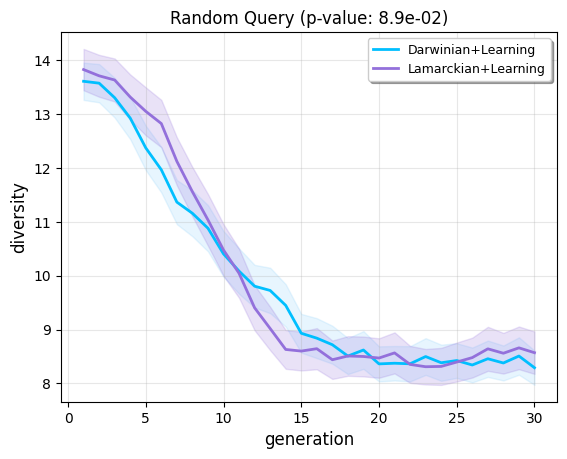}
         \caption{}
     \end{subfigure}
     \vspace{-5pt}
  \caption{Diversity over 30 generations (averaged over 10 runs) for Lamarckian system in purple and Darwinian system in blue. Subfigure (a) exhibits mean average diversity for robots produced with BFS, and Subfigure (b) is for Random Query.}
  \label{fig:diversity} 
\end{figure*}

Figure \ref{fig:diversity} demonstrates that initially, robots generated by BFS exhibit greater diversity compared to those generated by Random Query. Moreover, the morphological diversity of the Lamarckian system using BFS diminishes at a notably faster rate than the other three methods, indicating a convergence toward superior body designs at a faster pace. In the case of the Darwinian system, employing BFS led to a higher diversity value at the conclusion of the evolutionary process.

\begin{figure*}[h!] 
  \centering
     \begin{subfigure}[b]{0.49\textwidth}
         \centering
\includegraphics[width=0.9\textwidth]{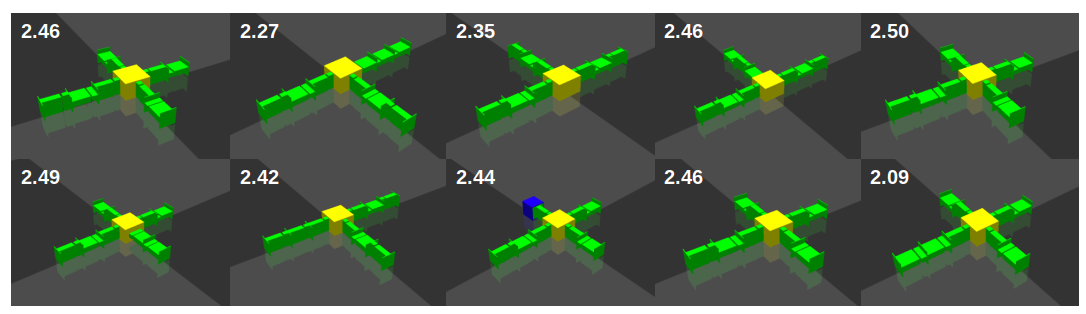}
         \caption{BFS (Darwinian system)}
     \end{subfigure}
     \hfill
     \begin{subfigure}[b]{0.49\textwidth}
         \centering
\includegraphics[width=0.9\textwidth]{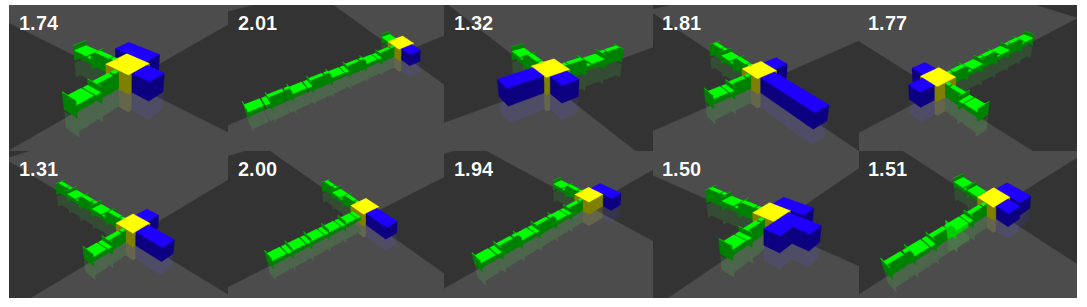}
         \caption{Random Query (Darwinian system)}
     \end{subfigure}

  \centering
     \begin{subfigure}[b]{0.49\textwidth}
         \centering
\includegraphics[width=0.9\textwidth]{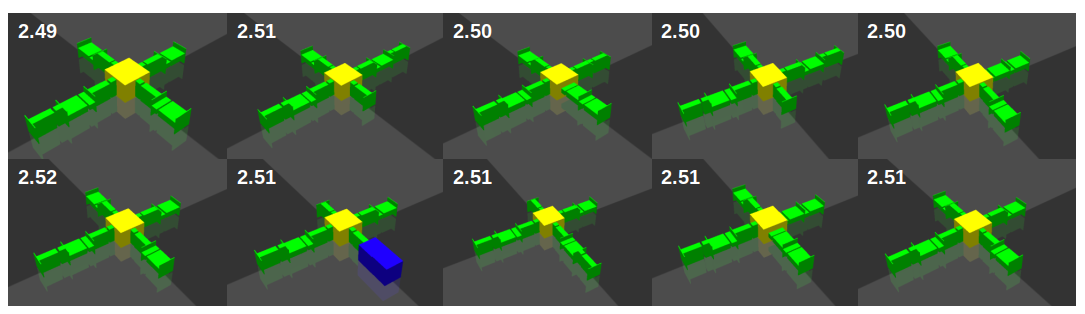}
         \caption{BFS (Lamarckian system)}
     \end{subfigure}
     \hfill
     \begin{subfigure}[b]{0.49\textwidth}
         \centering
\includegraphics[width=0.9\textwidth]{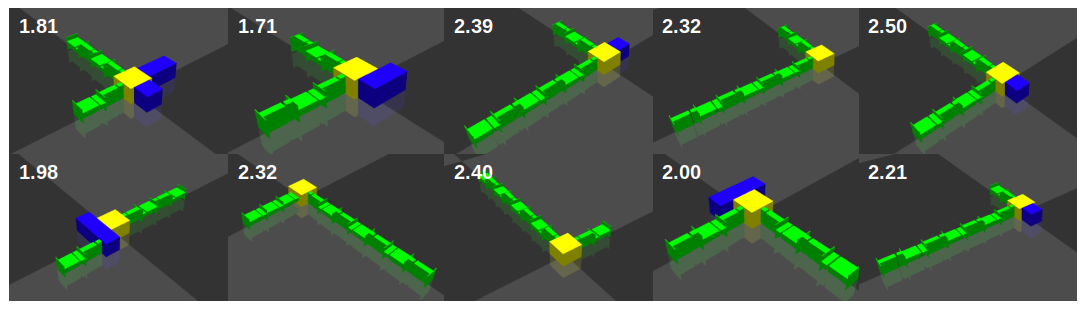}
         \caption{Random Query (Lamarckian system)}
     \end{subfigure}
  \caption{The 10 best robots produced by two query mechanisms with two evolution systems in each run with their fitnesses.}
  \label{fig:best5}
\end{figure*}




\subsubsection{Morphological traits}
We additionally examine the morphological characteristics of the robots, delving into eight specific traits (further information on the measurements can be found in [\cite{miras2018search}]).

\begin{figure*}[ht!]
\input{descriptors.tex}
\caption{Morphological traits over generations for both query mechanisms with two evolution systems. We present the progression of their means averaged over 10 runs for the entire population. Shaded regions denote a 95\% confidence interval.}
\label{fig:fitness_mean_descriptor} 
\end{figure*}
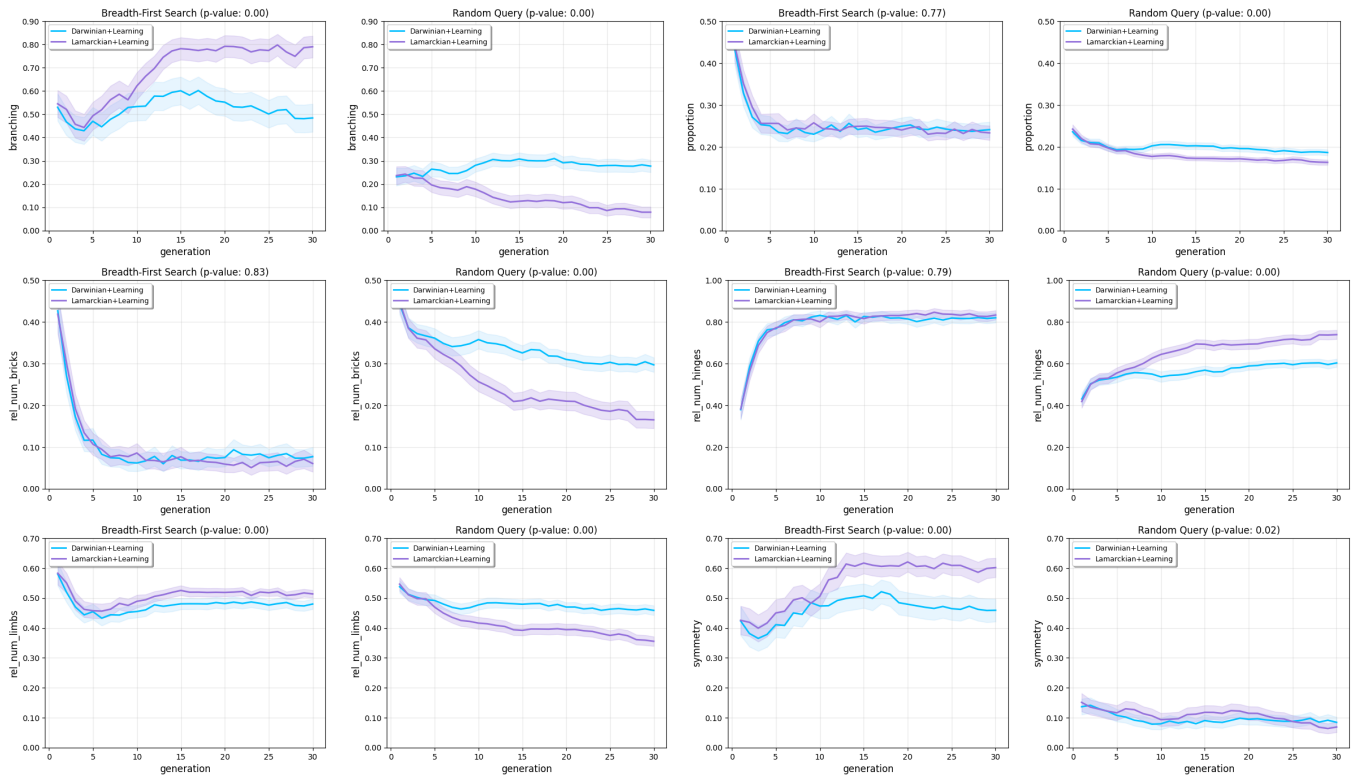

Figure \ref{fig:fitness_mean_descriptor} illustrates that the differences among robots generated by two evolutionary systems are notably larger when employing the Random Query method across all morphological traits, except for branching and symmetry, as opposed to using BFS.

Except for 'rel\_num\_bricks,' the values in all the other morphological traits from BFS are higher than those from Random Query. This means that robots produced by BFS are much more symmetrical, have more branching, more hinges, and fewer bricks compared to the ones produced by Random Query.
 
\begin{figure*}
    \centering
    \begin{subfigure}[b]{0.49\textwidth}
        \centering
        \includegraphics[width=0.94\textwidth]{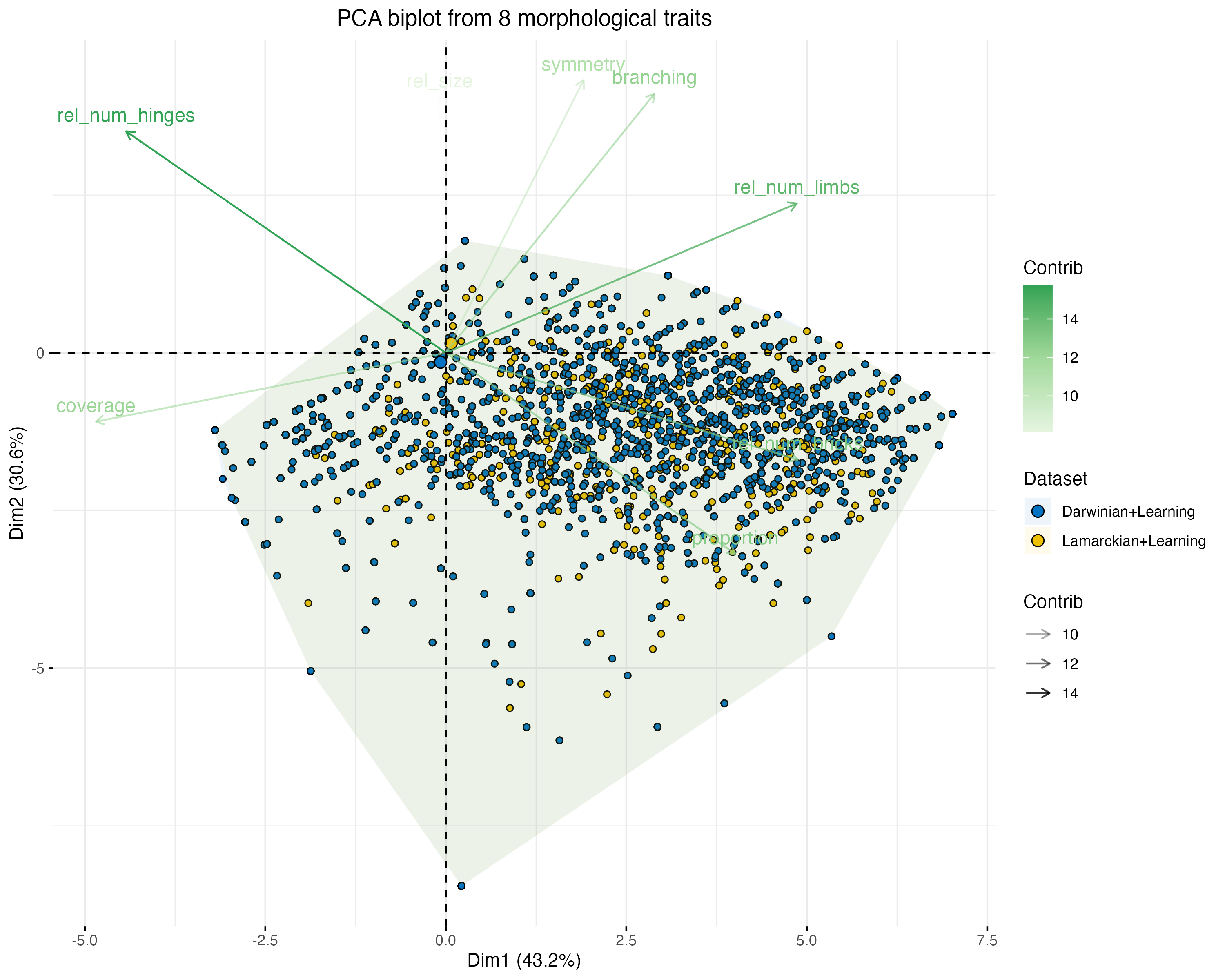}
     \caption{}
    \end{subfigure}
    \hfill
    \begin{subfigure}[b]{0.49\textwidth}  
        \centering 
        \includegraphics[width=0.94\textwidth]{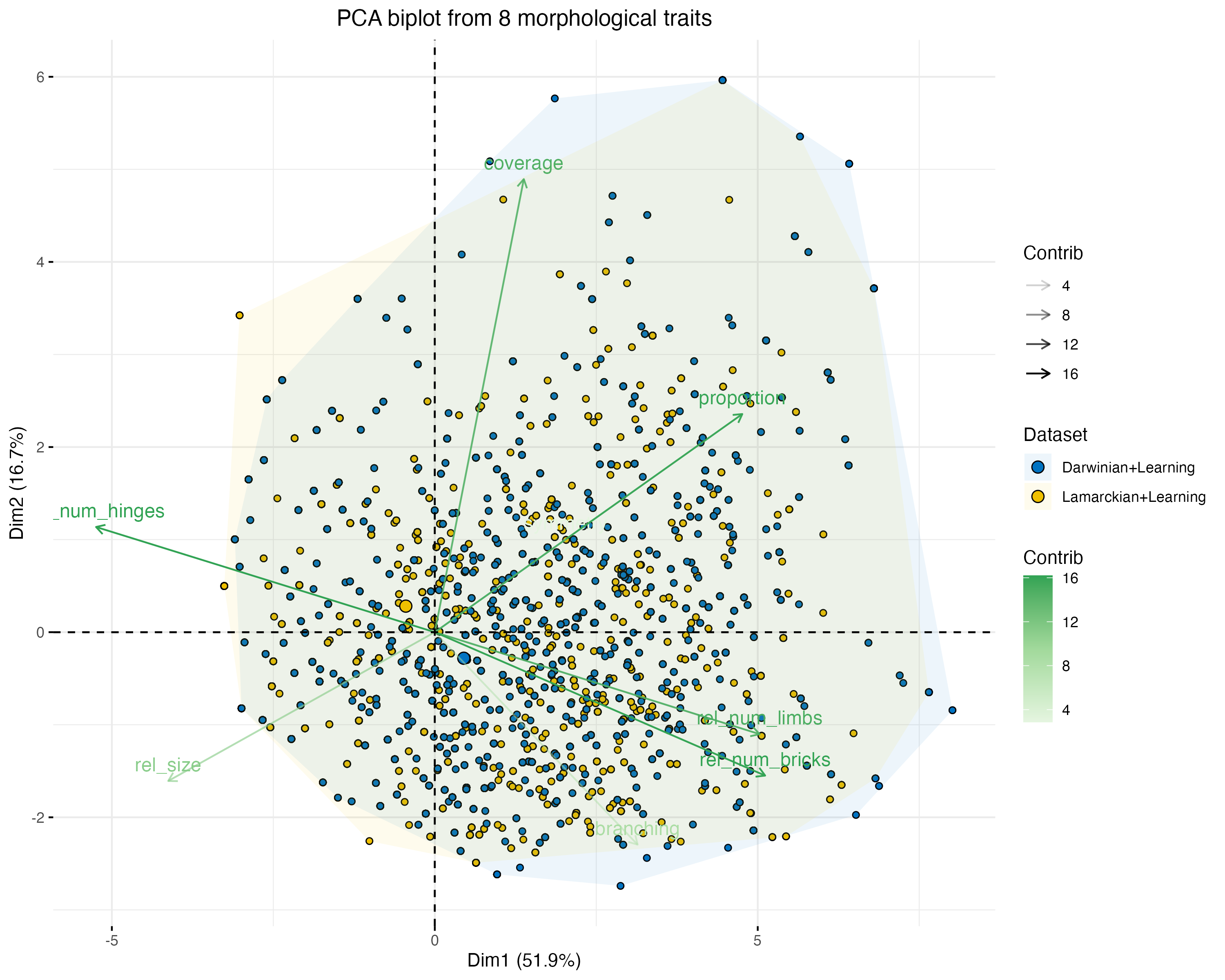}
        \caption{}
    \end{subfigure}
    
    \caption{Principal Component Analysis (PCA) biplots illustrate the distribution of samples based on eight morphological traits from two evolutionary systems for each search mechanism. Subfigure (a) displays datasets for BFS, while Subfigure (b) shows Random Query results. In each plot, every point represents a robot sample, and the axes represent the first two principal components (Dim1 and Dim2), which collectively explain the percentage of the total variance between Lamarckian and Darwinian system. Additionally, the biplot depicts the morphological traits as arrows, symbolizing their contributions to the principal components. Traits pointing in similar directions indicate co-regulation or similar expression patterns across the samples.}
    \label{fig:pca}
\end{figure*}

Furthermore, a PCA analysis (Figure \ref{fig:pca}) employing these identical eight traits reveals no difference in the morphologies generated by two evolutionary systems using BSF (subplot a). When employing the Random Query approach, there is a slight variation in the clustering circles (subplot b).

Hence, when applying the same query mechanism, the distinctions in the robots produced by the two evolutionary systems are marginal, whereas the differences in the robot bodies resulting from the two query mechanisms are considerable.

This is also supported by Figure \ref{fig:best5} which displays the 10 best robots produced by each method. The morphologies of the best-performing robots using BFS mainly converged into a "+" shape, while using the Random Query, the morphologies predominantly converge into an "L" shape, irrespective of the evolution system used. The best morphologies evolved by BFS from both evolution systems typically feature three or four limbs, primarily consisting of hinges with either no bricks or just one. In contrast, those generated through the Random Query method tend to have a relatively higher likelihood of containing one or two bricks and consist of only two limbs.

\section{Conclusions and Future Work}

In this research, we investigated the influence of two different query mechanisms used in genotype to phenotype mapping within two evolutionary robotics systems. Based on our analysis, we draw the following conclusions:

Firstly, the choice of query mechanism significantly affects the evolution and performance of modular robot bodies. Robots queried by BFS exhibited approximately 20\% better efficacy in solving the given task. Additionally, BFS in the Lamarckian system demonstrated superior efficiency, finding the best solution faster compared to Random Query.

Secondly, the query mechanism plays a crucial role in shaping the morphological intelligence of evolved robot bodies. Our experiments showed that morphological intelligence, measured as the ability of the body to facilitate learning in the brain, was significantly higher in robots produced by BFS. This highlights the importance of the query mechanism in determining the learning potential and adaptability of the evolved robot morphologies.

Furthermore, our analysis revealed that the query mechanism influenced the diversity and morphological traits of the evolved robot bodies. Robots produced by BFS exhibited higher diversity initially. In the Lamarckian system, it declines faster, converging to superior designs, while in the Darwinian system, BFS led to higher end-process diversity. Regarding morphological traits, for the same query mechanism, the distinctions in the robots produced by the two evolutionary systems are marginal, whereas the differences in the robot bodies resulting from the two query mechanisms are considerable.

In conclusion, BFS offers a systematic and deterministic approach, ensuring the exploration of every possible branch of the genotype tree. This results in increased stability and efficiency. On the contrary, the Random query approach, in theory, introduces variability that might lead to innovative body designs – the primary rationale behind our initial choice. However, our experimental results do not definitively showcase any discernible advantages. As we move forward, there is scope to explore alternative query mechanisms within various evolutionary frameworks.

\bibliographystyle{IEEEtran}
\bibliography{references} 

\end{document}

%% file: descriptors.tex
\begin{minipage}[c]{0.99\textwidth} 
        \begin{minipage}[t]{0.242\textwidth}
            \centering
            {\includegraphics[width=\textwidth]{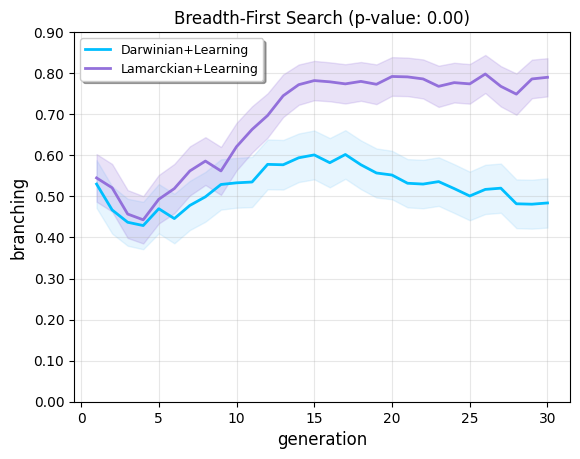}}
        \end{minipage}
        \hfill
        \begin{minipage}[t]{0.242\textwidth}
            \centering
            {\includegraphics[width=\textwidth]{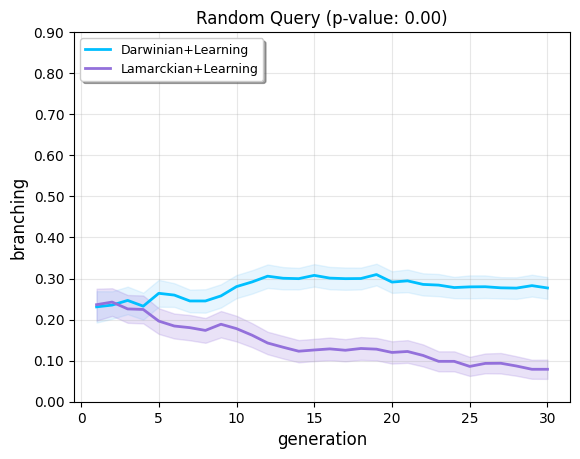}}
        \end{minipage}
        \hfill
        \begin{minipage}[t]{0.242\textwidth}
            \centering
            {\includegraphics[width=\textwidth]{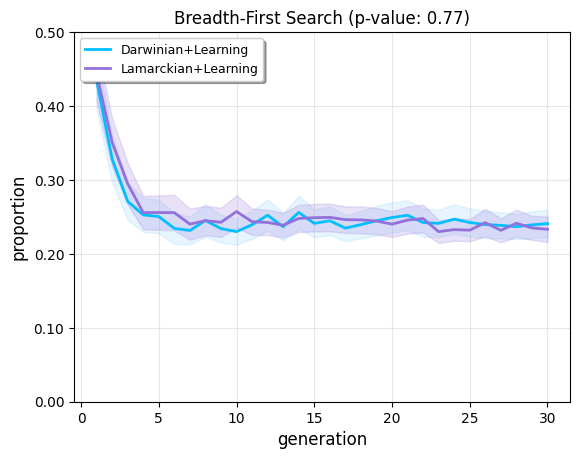}}
        \end{minipage}
        \hfill
        \begin{minipage}[t]{0.242\textwidth}
            \centering
            {\includegraphics[width=\textwidth]{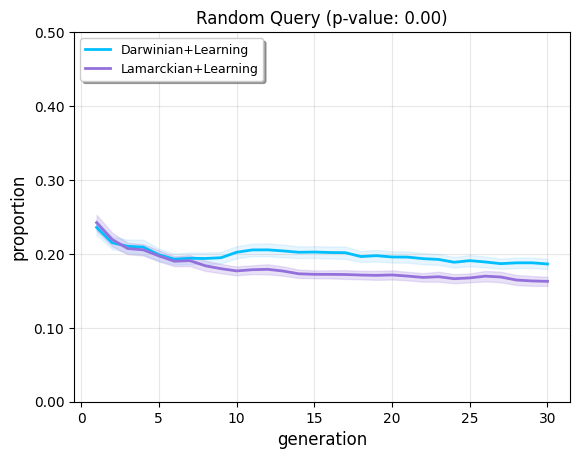}}
        \end{minipage}
        \hfill
\end{minipage}

\begin{minipage}[c]{0.99\textwidth}  
        \begin{minipage}[t]{0.242\textwidth}
            \centering
            {\includegraphics[width=\textwidth]{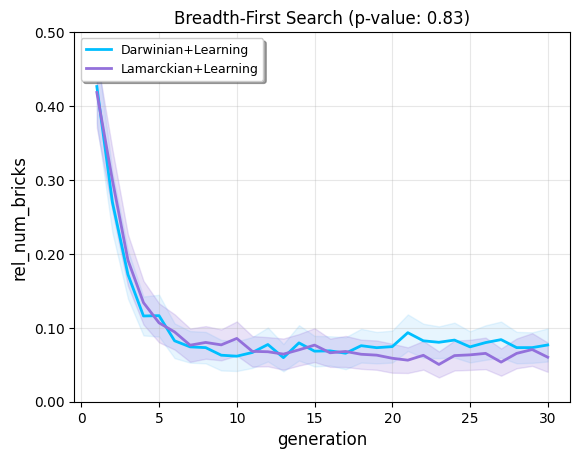}}
        \end{minipage}
        \hfill
        \begin{minipage}[t]{0.242\textwidth}
            \centering
            {\includegraphics[width=\textwidth]{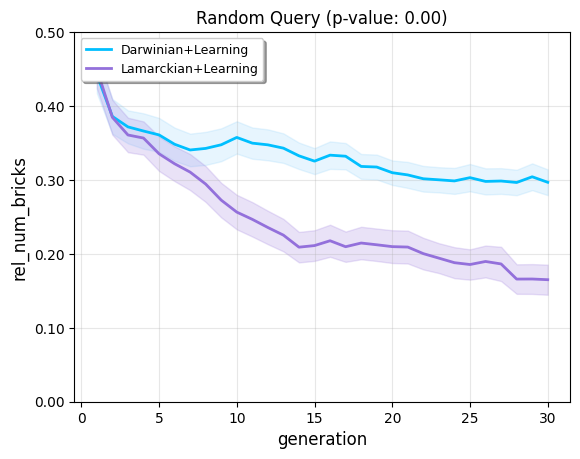}}
        \end{minipage}
        \hfill
        \begin{minipage}[t]{0.242\textwidth}
            \centering
            {\includegraphics[width=\textwidth]{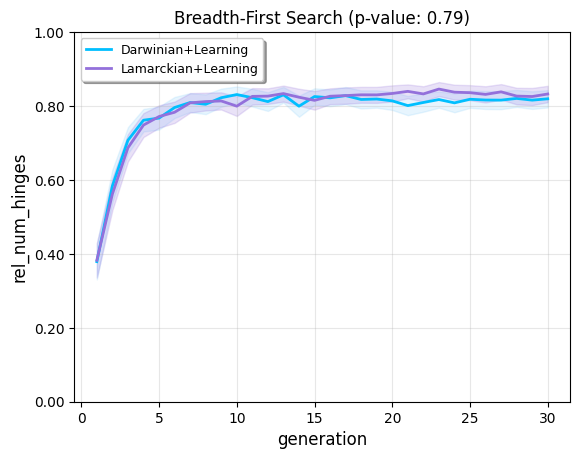}}
        \end{minipage}
        \hfill
        \begin{minipage}[t]{0.242\textwidth}
            \centering
            {\includegraphics[width=\textwidth]{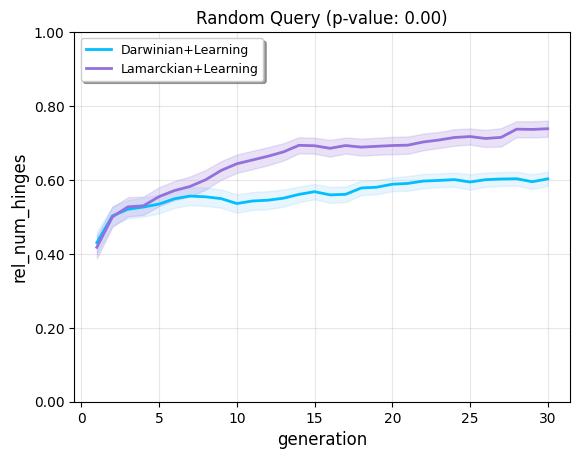}}
        \end{minipage}
\end{minipage}

\begin{minipage}[c]{0.99\textwidth}  
        \begin{minipage}[t]{0.242\textwidth}
            \centering
            {\includegraphics[width=\textwidth]{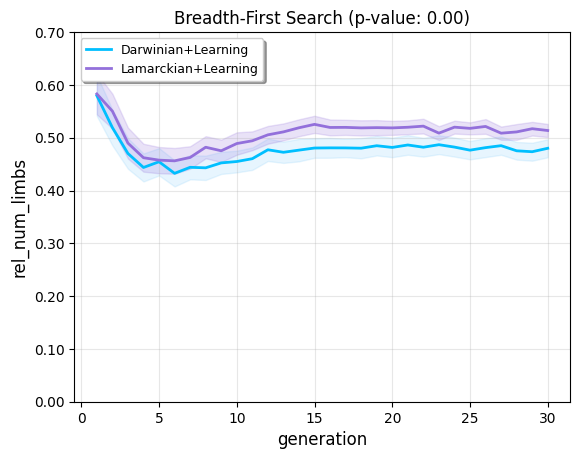}}
        \end{minipage}
        \hfill
        \begin{minipage}[t]{0.242\textwidth}
            \centering
            {\includegraphics[width=\textwidth]{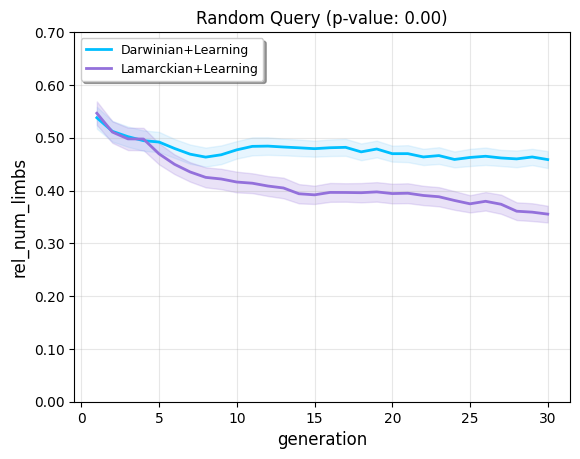}}
        \end{minipage}
        \hfill
        \begin{minipage}[t]{0.242\textwidth}
            \centering
            {\includegraphics[width=\textwidth]{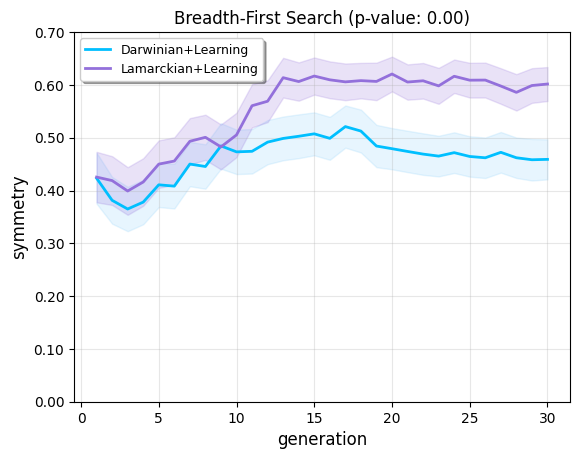}}
        \end{minipage}
        \hfill
        \begin{minipage}[t]{0.242\textwidth}
            \centering
            {\includegraphics[width=\textwidth]{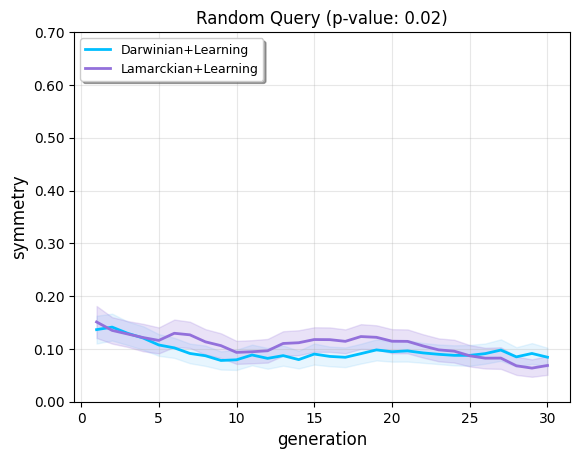}}
        \end{minipage}
\end{minipage}